\documentclass[11pt,a4paper]{article}
\usepackage[hyperref]{emnlp-ijcnlp-2019}
\usepackage{times}
\usepackage{latexsym}

\usepackage{url}
\usepackage{times}
\usepackage{soul}
\usepackage{graphicx}
\usepackage{amsmath}
\usepackage{booktabs}
\usepackage{algorithm}
\usepackage{algorithmic}
\usepackage{multirow}
\usepackage{amssymb}
\usepackage{xcolor}
\usepackage{enumitem}

\usepackage[symbol]{footmisc}

\aclfinalcopy 

\title{PaRe: A Paper-Reviewer Matching Approach Using a Common Topic Space}

 \author{Omer Anjum$^{*,1}$, Hongyu Gong$^{*,1}$, Suma Bhat$^{1}$, Jinjun Xiong$^{2}$, Wen-Mei Hwu$^{1}$ \\
  $^1$University of Illinois at Urbana-Champaign, USA \\
  $^2$ IBM Thomas J. Watson Research Center, USA \\
  \texttt{\{oanjum, hgong6, spbhat2, w-hwu\}@illinois.edu, jinjun@us.ibm.com}
  }

\date{}

\begin{document}
\maketitle

\begin{abstract}
Finding the right reviewers to assess the quality of conference submissions is a time consuming process for conference organizers. Given the importance of this step, various automated reviewer-paper matching solutions have been proposed to alleviate the burden. Prior approaches, including bag-of-words models and  probabilistic topic models have been inadequate to deal with the vocabulary mismatch and partial topic overlap between a paper submission and the reviewer's expertise.
Our approach, the common topic model,  jointly models the topics common to the submission and the reviewer's  profile while relying on abstract topic vectors. 
Experiments and insightful evaluations on two datasets demonstrate that the proposed method  achieves consistent improvements  compared to available state-of-the-art implementations of paper-reviewer matching.

\end{abstract}
\footnotetext{*Omer Anjum and Hongyu Gong have equal contribution.}

\section{Introduction}


The peer review mechanism constitutes the bedrock of today's  academic research landscape spanning submissions to conferences, journals, and funding bodies across numerous disciplines. Matching a  paper (or a proposal) to an expert in the topic presented in the paper requires the knowledge of diverse topics of both the submission as well as that of the reviewer's expertise in addition to knowing recent affiliations and co-authorship to resolve conflict of interest. Considering the scale of current conference submissions,  performing the task of paper-reviewer matching manually incurs significant overheads to the program committee (PC) and calls for automating the process. Faced with record number of paper submissions  essentially interdisciplinary in nature, the inadequacy of available reviewer matching systems to scale to the current needs is being expressed by many conference program committees.  
It is also notable that the approaches to address the challenges seem ad-hoc and non-scalable, as described in a few of the PC blogs; ``Looking at the abstracts for many of the submissions it also quickly became clear that there was disparity in how authors chose topic keywords for submissions with many only using a single keyword and others using over half a dozen keywords. As such relying on the keywords for submissions became difficult. The combined effect of these problems made any automatic or semi-automatic assignment using HotCRP suboptimal...So, we chose to hand assign the papers.'' \cite{sigarchBlog}, and again in ``Our plan was to rely on the Toronto Paper Matching System (TPMS) in allocating papers to reviewers. Unfortunately, this system didn’t prove as useful as we had hoped for (it requires more extensive reviewer profiles for optimal performance than what we had available) and the work had to rely largely on the manual effort...'' \cite{aclBlog}.  Noting the urgent need to advance research to address this problem,  we study this challenge of matching a paper  with a reviewer from a list of potential reviewers  for the purpose of assessing the quality of the submission.


Aside from  the long precedence of research in the related area of expertise retrieval -- that of expert finding and expert profiling  \cite{balog2012expertise}, several recent attempts have been made to automate the process \cite{price_flach_2017}. 
These include, the Toronto paper matching system  \cite{tpms}, the IEEE INFOCOM review assignment system \cite{li2016new}, and the online reviewer recommendation system \cite{Qian2018}. Central to  these systems is a module that performs the paper-reviewer assignment, which can be broken down into its matching and constraint satisfaction constituents. The constraint satisfaction component  typically handles the constraints that each paper be reviewed by at least a few reviewers, each reviewer be assigned no more than a few papers,  and thaat reviewers not be assigned papers for which they have a conflict of interest. A second constituent is that of finding a reviewer from a list of reviewers based on the relevance of the person's expertise  with the topic of the submission. This latter aspect will be the focus of our study.

Available approaches to solve this matching problem can be broadly classified into the following categories  \cite{price_flach_2017}: a) Feature-based matching, where a set of topic keywords are collected for each paper and each reviewer. The reviewers are then ranked in order of the number of keyword matches with the paper; b) Automatic feature construction with profile-based matching, where the relevance is decided by building automatic topic representations of both papers and reviewers; c) Bidding, a more  recent method, involves giving the reviewers access to all the papers and asking them to bid on papers of their choice. The approaches used in this study are of the profile-based matching kind, where we rely on  the use of abstract topic vectors and word embeddings \cite{Mikolov2013} to derive the semantic representations of the paper  and the expertise area of the reviewer. This is a departure from the bag-of-words approach taken in related prior approaches, e.g. \cite{tpms}, relying on automatic topic extraction using keywords -- a ranked list of terms taken from the paper and the reviewer's profile.

In general, we assume that a reviewer can be represented by a collection of the abstracts of her past publications (termed as the reviewer's profile) and a  submission by its abstract. While attempting to match the paper with the reviewer via their profile representations, the  obvious difference in the document lengths gives rise to a mismatch  due to the  small overlap in  vocabulary and a consequent scarcity of shared contexts of these overlapping terms. This is because, while past publications of a reviewer may be sufficient to provide a reasonable context for a topical word,  a submission abstract provides a very limited context for that topical word.



To alleviate this problem of mismatched `profiles', we use the idea of a shared topic space between the submission and the reviewer's profile. In our experiments we compare our approach to match the profiles using abstract vectors with that using the hidden topic model (also a set of abstract topic vectors) \cite{gong2018document}. The two approaches primarily differ in the way the shared topic space is constructed, which we describe in Section~\ref{sec:model}. We also include other baseline comparisons where the matching is done on the basis of common topic words (keywords) and word- or document-embeddings.

This study makes the following contributions: \\
(1) Instead of relying on a collection of topic words (keywords chosen by the authors or experts), our approach relies on abstract topic vectors to represent  the common topics shared by the submission and the reviewer. \\
(2) We propose a model that outperforms state-of-the-art approaches in the task of paper-reviewer matching on a benchmark dataset \cite{Mimno2007}. Additionally, a field evaluation of our approach performed by the program committee of a tier-1 conference showed that it was highly useful.

    
\section{Related Work}

The paper-reviewer matching task lays the basis for the peer review process ubiquitous in academic conferences and journals. 
Existing automatic approaches can be broadly categorized into the following types according to the type of models for comparing documents:   \emph{feature-based models}, \emph{probabilistic models}, \emph{embedding-based models}, \emph{graph models} and \emph{neural network models}.

\noindent\textbf{Feature-based models}. A list of keywords which summarizes the topics of a submission is used as informative features in the matching process \cite{Dumais1992,Basu99}. Automatic extraction of these features achieves higher efficiency and one commonly-used feature is a bag-of-words  weighted by the words' TF-IDF scores  \cite{LDA2,Tang2010,li2016new,Nguyen2018}. 

\noindent\textbf{Probabilistic models}. The Latent Dirichlet Allocation (LDA) model is the most commonly used probabilistic model in expertise matching, where each topic is represented as a distribution over a given vocabulary and each document is a mixture of hidden topics \cite{LDA}. The popular Toronto Paper Matching System (TPMS) \cite{tpms} uses LDA to generate the similarity score between a reviewer and a submission \cite{li2016new}. One limitation of LDA is that it does not make use of the potential semantic relatedness between words in a topic because of its assumption that words are generated independently \cite{xie2015incorporating}. Variants of LDA have been proposed to incorporate notions of semantic coherence  for more effective topic modeling  \cite{hu-tsujii-2016-latent, Gaussian-LDA-Word-Embeddings, Xun:2017}. Beyond having probabilistic models for topics, \citeauthor{jin2017integrating} sought to capture the temporal changes of reviewer interest as well as the stability of their interest trend with probabilistic modeling \cite{jin2017integrating}. 

In addition to their inherent limiting assumptions such as independence of semantically related words, probabilistic models, including LDA, require a large corpus to accurately identify the topics and topic distribution in each document, which can be problematic when applied to short documents, such as abstracts.

\noindent\textbf{Embedding-based models}. 
Latent Semantic Indexing (LSI) proposes to represent a document as a single dense vector \cite{deerwester1990indexing}. The documents corresponding to reviewers and submissions can thus be transformed into their respective vector representations. The relevance of a reviewer to a given submission would  then be measured using a distance metric in the vector space, such as the cosine similarity. Other approaches have used word or document embeddings as document representations in order to compare two documents.

\citeauthor{kou2015topic} derived topic vectors by treating each topic as a distribution of words \cite{kou2015topic}.  In comparison, the key improvement in our work is that the topics are derived based on word embeddings instead of word distributions. Moreover, we derive common topics for each submission-reviewer pair, and as a result, the topics can vary from pair to pair.

Another  approach to capture similarity between documents is by the use of the Word Mover's Distance (WMD). It relies on the alignment of word pairs from two texts, and the textual dissimilarity is measured as the total distance between the vectors of the word pairs \cite{WMD}. More recently, a hidden topic model has been used to compare two documents via extracted abstract topic vectors, which showed a strong performance in comparing document for semantic similarity \cite{gong2018document}.

Extending the models in this category, we propose the common topic model. Similar to the hidden topic model, we extract topic vectors using word embeddings and match documents at the topic level.
The hidden topic model extracts topics purely relying on the reviewer profile, so the  topic vectors can be regarded as a summary of the reviewers' research interest. In contrast, the common topic vectors are selected based on the knowledge of both the submission and the reviewer's profile, which are expected to capture the topical overlap between the two.  As we will show in the qualitative evaluation, the hidden topic model is likely to miss some important topics when a reviewer has broad research interests, resulting in an underestimation of the paper-reviewer relevance. The common topic model is able to overcome this limitation by extracting topics with reference to submissions.

\noindent\textbf{Graph models}. All of the models mentioned above only assume access to the texts of submissions and reviewers' publications. Some works also make use of external information such as coauthorship to improve the matching performance. For instance, \citeauthor{RWR} capture academic connections between reviewers using a graph model, and show that such information improves the matching quality. Each node in their graph model represents a reviewer \cite{RWR}. There is an edge between two nodes if the corresponding reviewers have co-authored papers, and the edge weight is the number of publications. This work also uses LDA to measure the similarity between the submission and the reviewer.

\noindent\textbf{Neural network models}. Dense vectors are learned by neural networks as the semantic representation of documents \cite{socher2011parsing,le2014distributed,lin2015hierarchical,lau2016empirical}. When it comes to the task of expertise matching, the reviewer-submission relevance can thus be measured by the similarity of the vector representations of their textual descriptions.

\section{System Overview}
The different stages of our system, together called PaRe, are briefly explained as below:\\ 
\textit{Data collection}. At this stage, we collect previous publications from one or more tier-1 conferences in the same domain as the one to which reviewer-submission matching is applied. This data is used to create our pool of candidate reviewers and domain knowledge of the research area.
The source of the data is Microsoft Academic Graph (MSG) \cite{MAG}. All the abstracts of a reviewer are concatenated as one document, which is then used to profile the reviewer. Reviewers' profiles reflect their research topics, which are later used in the reviewer-submission matching process.  \\
\textit{Data processing}. Since our proposed model is based on word embeddings, we pre-train embeddings using CBOW model of word2vec on the collected publications \cite{w2v}. The dense word representations are intended to capture domain-specific lexical semantics. The data collection and processing are detailed in Section \ref{sec:exp}. \\
\textit{Reviewer-submission matching}. A common topic modeling approach is proposed in this work to match reviewers with submissions.
The model compares the abstracts of submissions and reviewers' past abstracts during the matching process to decide the reviewer-submission relevance by finding their common research topics. The algorithm is described in Section \ref{sec:model}.

\section{Modeling}
\label{sec:model}
For the purpose of our study, we consider a reviewer's profile to be the concatenation of the abstracts from their previous publications.
Let $m$ be the number of words in the reviewer's profile and let the \emph{normalized} word embeddings of these words be stacked as a reviewer matrix $\bf{R}\in\mathbb{R}^{d\times m}$, where $d$ is the embedding dimension. Since the embedding is normalized, we have $\lVert\bf{R}_{i}\rVert_{2}=1$ for each column in $\bf{R}$.
Next suppose that the submission is represented by an $n-$word sequence of its abstract. Similar to the case of the reviewer, we stack its normalized embeddings as a submission matrix $\bf{S}\in\mathbb{R}^{d\times n}$. Also we have for each column $\lVert \bf{S}_{j}\rVert_{2}=1, \forall 1\leq j\leq n$.

\noindent\textbf{Common topic selection}.  Inspired by the compositionality of embeddings \cite{gong2017geometry} and the hidden topic model in document matching \cite{gong2018document}, our intention is to extract topics from reviewer profiles and submissions to summarize their topical overlap. We would like to remind the reader that the topics extracted  are neither words nor distributions, but only abstractions and constitute a set of numeric vectors that do not necessarily have a textual representation. To establish the connection between topics, reviewer profiles and submissions, we assume that the topic vectors can be written as a linear combination of the embeddings of component words in either the reviewer profiles or the submissions. This assumption is supported by the geometric property of word embeddings that the weighted sum of the component word embeddings have been shown to be a robust and efficient representation of sentences and documents \cite{Mikolov2013}. Intuitively, the extracted common topics would be highly correlated with the subset of the words in the reviewer profile or that of the submission in terms of semantic similarity.

Let both the reviewer and the submission have $K$ research topics, with each topic represented by a $d-$dimensional vector. This vector is an abstract topic vector and does not necessarily correspond to a specific word or a word distribution as in LDA \cite{LDA}. Suppose that these topic vectors of the reviewer are stacked as a matrix ${\bf P}\in \mathbb{R}^{d\times K}$, and those of the  submission as ${\bf Q}\in\mathbb{R}^{d\times K}$. Therefore, these matrices can be represented as a linear combination of the underlying word vectors. 
\begin{align}
\nonumber
{\bf P} &= {\bf Ra}, \\
\label{eq:linear_combination}
{\bf Q} &= {\bf Sb},
\end{align}
where ${\bf a} \in\mathbb{R}^{m\times K}, {\bf b}\in \mathbb{R}^{n\times K}$ are the coefficients in the linear combinations.

Our goal is to find the common topics shared by a given reviewer and a given submission, to account for  the overlap of their research areas. We consider a pair of topics from a reviewer and a submission respectively to constitute a pair of common topics if they are semantically similar. For example, if the reviewer's research areas are \emph{machine learning} and \emph{theory of computation}, and the submission is about \emph{classification} in \emph{natural language processing}, then (\textit{machine learning, classification}) can be regarded as a pair of common topics, while the other pairs corresponding to the areas \textit{theory of computation} and \textit{natural language processing} are much less similar. We used cosine similarity to measure the semantic similarity of two topic vectors \footnote{We leave it to future work to experiment with other useful measures of semantic similarity.}. The similarity $\text{sim}({\bf P}_{k}, {\bf Q}_{k})$ between reviewer topic ${\bf P}_{k}$ and submission topic ${\bf Q}_{k}$ is shown below.
\begin{align}
    \text{sim}({\bf P}_{k}, {\bf Q}_{k}) = \frac{{\bf P}_{k}^{T}{\bf Q}_{k}}{\lVert{\bf P}_{k}\rVert\cdot \lVert {\bf Q}_{k}\rVert}.
\end{align}
For $K$ pairs of topic vectors $\{{\bf P_{k}}, {\bf Q_{k}}\}_{k=1}^{K}$, their similarity is the sum of the pairwise similarities:
\begin{align}
\text{sim}({\bf P}, {\bf Q}) = \sum\limits_{k=1}^{K}\text{sim}({\bf P}_{k}, {\bf Q}_{k}).
\end{align}

This in turn translates to identifying the common research topics between the reviewer and the submission, i.e., we need to find $K$ such pairs of topics that have the maximum similarity. Based on our discussions above, the approach of common topic extraction can be formulated as an optimization problem:
\begin{align}
\nonumber
&\max\limits_{\mathbf{a},\mathbf{b}} \text{sim}({\bf P}, {\bf Q}) \\
\nonumber
\text{s.t.}\quad &\mathbf{P} = \mathbf{Ra}, \\
\nonumber
&\mathbf{Q} = \mathbf{Sb},\\
\label{eq:common_topic}  
&\mathbf{P}^{T}\mathbf{P}=\mathbf{Q}^{T}\mathbf{Q}=\mathbf{I}
\end{align}

The first two constraints are based on the linear assumption shown in Eq.~\ref{eq:linear_combination}.
Without loss of generality, we add the third constraint that the topic vectors are orthogonal as shown in Eq.~\ref{eq:common_topic} to avoid generating multiple similar topic vectors. The closed-form solution to this optimization problem can be derived via singular value decomposition on the correlation matrix of $\bf{R}$ and $\bf{S}$  \cite{wegelin2000survey}.

Let topic vectors $\bf{P}^{*}$ and $\bf{Q}^{*}$ be the optimal solution, both describing the common topics shared by the reviewer and the submission. In the following discussions, we use $\bf{P}^{*}$ as the common topic vectors.

\noindent\textbf{Common topic scoring}. To further quantify the reviewer-submission relevance, we need to evaluate how significant these common topics are for the reviewer and the submission respectively. Reusing the example where a reviewer's area are \textit{machine learning} and \textit{theory of computation},  we know that \textit{machine learning} is the common topic between the reviewer and the submission. If the topic of \textit{machine learning} were only a small part of the reviewer's publications, the reviewer may not be a good match for the submission since reviewer is more of an expert  in \textit{theory of computation} than in \textit{machine learning}.

To evaluate how well the topics reflect a reviewer's expertise, we define the importance of common topics ${\bf P}^{*}$ for both the reviewer and the submission. Consider the vector of the $i-$th word in the reviewer's profile, ${\bf R}_{i}$, and the $k$-th topic vector ${\bf P}_{k}^{*}$. The relevance between ${\bf{R}}_{i}$ and ${\bf P}_{k}^{*}$ is defined as the their squared cosine similarity.
\begin{align}
\text{rel}({\bf R}_{i}, {\bf P}_{k}^{*}) = \text{cos}^{2}({\bf R}_{i}, {\bf P}_{k}^{*}) = ({\bf R}_{i}^{T}{\bf P}_{k}^{*})^{2}.
\end{align}

Note that we do not use cosine similarity as is, since  $\mathbf{R}_{i}$ and $\mathbf{P}_{k}^{*}$ might be negatively correlated and the cosine similarity can be negative. Instead, we use the square of the cosine similarity  to reflect the strength of their correlation.


The relevance between word ${\bf R}_{i}$ and a set of topic vectors ${\bf P}^{*}$ is defined as the sum of the relevance between the word and each topic vector.
\begin{align}
\text{rel}({\bf R}_{i}, {\bf P}^{*}) = \sum\limits_{k=1}^{K}\text{rel}({\bf R}_{i}, {\bf P}_{k}^{*}).
\end{align}
We can think of word vector ${\bf R}_{i}$ to be projected along the $K$ dimensions of a linear subspace spanned by topic vectors in $\bf{P}^{*}$. If ${\bf R}_{i}$ lies in this linear subspace, then it can be represented as a linear combination of the topic vectors. In this case, $\text{rel}({\bf R}_{i}, {\bf P})$ achieves the maximum of $1$. If the word vector is orthogonal to all topic vectors in ${\bf P}^{*}$, the relevance results in the minimum relevance of $0$. Thus, the range of $\text{rel}({\bf R}_{i}, {\bf P})$ is from 0 to 1.

Furthermore, we define the relevance between the reviewer and the topics as the average of the relevance between the words and the topics.
\begin{align}
\text{rel}({\bf R}, {\bf P}^{*}) = \frac{1}{m}\sum\limits_{i=1}^{m}\text{rel}({\bf R}_{i}, {\bf P}^{*}).
\end{align}
The reviewer-topic relevance $\text{rel}({\bf R}, {\bf P})$ also ranges from 0 to 1.

Similarly, we measure the relevance between a submission and a set of common topics, $\text{rel}({\bf S}, {\bf P}^{*})$ by measuring the relevance between words in the submission and common topics. The submission-topic relevance reflects the importance of the common topics for a submission. We define the reviewer-submission matching score as a harmonic mean (f-measure) of the reviewer-topic and submission-topic relevance \cite{powers2015f}. 

\begin{align}
\label{eq:score}
\text{rel}({\bf R}, {\bf S}) = \frac{2\cdot \text{rel}({\bf R}, {\bf P}^{*})\cdot \text{rel}({\bf S}, {\bf P}^{*})}{\text{rel}({\bf R}, {\bf P}^{*}) + \text{rel}({\bf S}, {\bf P}^{*})}.
\end{align}


The reviewer-submission relevance is high when the common topic vectors ${\bf P}^{*}$ are highly relevant to both the reviewer and the submission. It indicates that the submission has a substantial overlap with reviewer's research area, and that the reviewer is considered to be a good match for the submission.

\begin{table*}[!h]
\centering
\begin{tabular}{|c|c|c|c|c|c|l|l|}
\hline
\textbf{Method}                     & \textbf{Number of Topics} & \textbf{P@5}    & \textbf{P@10}   & \textbf{P@5}    & \textbf{P@10}   & \textbf{P@5}    & \textbf{P@10}           \\ \hline
\multicolumn{1}{|l|}{}              & \textbf{}                 & \multicolumn{2}{c|}{\textbf{GT1}} & \multicolumn{2}{c|}{\textbf{GT2}} & \multicolumn{2}{c|}{\textbf{GT3}} \\ \hline
\multirow{3}{*}{\textbf{Common Topic Model}} & 5                         & 53.7            & 43.0            & 58.5            & 49.1            & 63.7            & \textbf{55.8}            \\ \cline{2-8} 
                                    & 10                        & \textbf{56.6}            & \textbf{44.6}            & \textbf{63.2}            & \textbf{50.4}            & \textbf{67.2}            & 55.2            \\ \cline{2-8} 
                                    & 20                        & 54.4            & 43.4            & 59.2            & 49.5            & 63.6            & 54.7            \\ \hline
\multirow{3}{*}{\textbf{Hidden Topic Model}} & 10                        & 43.4            & 41.1            & 46.4            & 45.4            & 61.8            & 46.3            \\ \cline{2-8} 
                                    & 20                        & 51.3            & 43.4            & 58.4            & 49.0            & 63.6            & 53.6            \\ \cline{2-8} 
                                    & 30                        & 47.5            & 40.0            & 49.6            & 44.0            & 56.3            & 49.4            \\ \hline
\textbf{APT200}     & 200          & 41.18       & 29.71      &      -      &       -      &      -      &      -       \\ \hline
\textbf{Single Doc} &      -        & 44.71       & 27.35      &       -     &       -      &      -      &       -      \\ \hline
\multirow{3}{*}{\textbf{LDA}}                & 50                        & 41.3            & 38.4            & 51.2            & 45.0            & 55.4            & 51.5            \\ \cline{2-8} 
                                    & 200                       & 47.5            & 37.3            & 53.6            & 43.6            & 50.9            & 50.0            \\ \cline{2-8} 
                                    & 300                       & 46.2            & 38.8            & 52.0            & 46.8            & 50.0            & 45.2            \\ \hline 
\textbf{HDP}        &      -        & 45.5          & 38.0      &        48.0    &       44.5      &      55.4      &      50.0       \\ \hline
\textbf{RWR}        &     -         & 45.3        & 43         &     -       &        -     &      -      &        -     \\ \hline
\textbf{Doc2Vec}        &      -        & 51.7        & 41.1       &     59.2      &     46.8        &    64.5        &     51.1        \\ \hline
\textbf{WMD}        &      -        & 36.1        & 32.4       &     42.2       &    36.8         &   46.8         &    41.8         \\ \hline
\end{tabular}
\caption{The mean precision of different baselines with optimal hyperparamters on the NIPS dataset. A reviewer is classified as relevant with a TREC score $\geq 2$.}
\label{tab:result1}
\end{table*}

\section{Experiments and Results}
\label{sec:exp}
In this section, we empirically compare our proposed common topic model approach against a variety of models in the task of expertise matching.

\subsection{Dataset}
For our experiments, we use the two datasets described below. 

\noindent \textbf{NIPS dataset}. This is a benchmark dataset described in \citep{Mimno2007} and commonly used in the evaluation of expertise matching. It consists of $148$ NIPS papers accepted in 2006 and abstracts from the publications of $364$ reviewers.  It includes annotations from $9$ annotators on the relevance of $650$ reviewer-paper pairs. Each pair is rated on a scale from 0 to 3. Here ``0'' means irrelevant, ``1'' means slightly relevant, ``2'' means relevant and ``3'' means very relevant.

\noindent \textbf{A new dataset.}  Our proposed paper-reviewer matching system is applied to a tier-1 conference in the area of computer architecture. We created a new dataset for the evaluation of expertise matching from the submissions to this conference. We first collected a pool of $2284$ candidate reviewers with publications in top conferences of computer architecture. A reviewer selection policy was adopted by the conference program committee to select reviewers still active in relevant areas. Reviewers were excluded if \\
1) they started publishing 40 years ago, but had no publications for the last ten years; \\
2) they did not have publications for the last ten years and have fewer than three papers before that. \\

The publications of these reviewers were collected from Microsoft Academic Graph \cite{MAG}. Each reviewer had at least one publication, and some reviewers had as many as 34 publications. Again the abstracts were used as reviewers' profile. 

We then used our proposed common topic model to assist the program committee of the conference on computer architecture, and recommended most relevant reviewers to all submissions in the conference. We randomly selected $20$ submissions and with the help of the committee, we collected feedbacks from 33 reviewers on their relevance to the assigned submissions. These 33 reviewers were among the top reviewers recommended by our system for each of 20 submissions. The relevance was rated on a scale from 1 to 5, where a score of ``1'' meant that the paper was not relevant at all, ``2'' meant that the reviewer had passing familiarity with the topic of the submission, ``3'' meant that the reviewer knew the material well, ``4'' meant that the reviewer had a deep understanding of the submission, and ``5'' means that the reviewer was a champion candidate for the submission. 

\subsection{Baselines}
We include previous approaches to paper-reviewer matching as our baselines.
\begin{itemize}[leftmargin=*]
\setlength{\itemsep}{1pt}
\setlength{\parskip}{0pt}
\setlength{\parsep}{0pt}  
\item \textbf{APT 200}. Author-Person-Topic \cite{Mimno2007} is a generative probabilistic topic model which groups documents of an author into different clusters with the author's topic distribution. Clusters represent different areas of a reviewer's research.   
\item \textbf{Single Doc}. The Single Doc model is a probabilistic model which takes the idea of language modeling and estimates the likelihood that a submission is assigned to a reviewer given the reviewer's previous works \cite{Mimno2007}.
\item \textbf{Latent Dirichlet Allocation (LDA)}: LDA and its variants are the most popular topic models in expertise matching systems \cite{LDA}. LDA models assume that each document is a mixture of topics where each topic is a multinomial distribution over the words. 
\item \textbf{Hierarchical Dirichlet Process (HDP)}.  HDP model is an extension of LDA \cite{hdp}. It is a non-parametric mixed-membership Bayesian model with variable number of topics. It is effective in choosing the number of topics to characterize a given corpus.
\item \textbf{Random Walk with Restart (RWR)}. RWR is a graph model with sparsity constraints in expertise matching \cite{RWR}. It relies on LDA to capture reviewer-submission relevance and also takes diversity into consideration in the matching process.
\item \textbf{Word Mover's Distance (WMD)}. WMD is a distance metric between two documents on the basis of pre-trained word embeddings \cite{WMD}. It calculates the dissmilarity between two documents, which is measured by the embedding distance of aligned words in these documents.
\item \textbf{Hidden Topic Model}. This model proposes to learn hidden topic vectors to measure document similarity based on word embeddings \cite{gong2018document}. 
\item \textbf{Doc2Vec}. Doc2Vec is a neural network model which trains document embeddings to predict component words in the documents \cite{le2014distributed}. In expertise matching, the Doc2Vec model is pre-trained on the corpus consisting of reviewers' previous publications. We use the trained model to generate representations for reviewers and submissions respectively. The reviewer-submission relevance is quantified by the cosine similarity of their embeddings.
\end{itemize}

\noindent\textbf{Setting}. Since our model relies on word embeddings, we pre-train embeddings on all papers published in the NIPS conference until 2017 for the matching task in NIPS dataset. Similarly for our new dataset, we collected a corpus of publications until 2018 from top computer architecture conferences for embedding training. The embedding dimension was set as 100, and these word embeddings were also used in two embedding-based baselines: word mover's distance and hidden topic model. For a fair comparison, the corpora used for word embedding training were also used to train Doc2Vec model to generate document embeddings.


\subsection{Results on NIPS Data}
The NIPS dataset provides ground truth relevance for reviewer-submission pairs, and the relevance scales from 0 to 3. 
A score of 0 is assigned when that the reviewer is considered to be \textit{not relevant} and a score of 3 is assigned when the reviewer is considered to be \textit{highly relevant}. We set a relevance threshold of $2$, and considered reviewers with a score equal to or higher than this threshold to be relevant reviewers to the given submission. In our matching system, we sorted reviewers in decreasing order of the predicted relevance score for a given submission.

\noindent\textbf{Evaluation Metric}. Precision at k (P@k) is a commonly-used ranking metric on NIPS dataset. P@k is defined to be the percentage of relevant reviewers in the top-k recommendations made by the model to a submission. It is likely that the top-k recommendations made by the model contain reviewers whose relevance information is not available in the ground truth. To address this issue, we first discard reviewers that do not have a relevance information prior to calculating P@k. In our experiments, we set k to be 5 and 10. We report the average P@k over all submissions in Table \ref{tab:result1}. 


We note that not all submissions in NIPS dataset have the same number of relevant reviewers and a failure to account for this  discrepancy would negatively impact the performance of a system. For example, a submission with only one relevant reviewer would result in a P@5 no higher than 20\% for any model. In order to take this discrepancy into consideration,  we report the performance only on submissions with at least two relevant reviewers in the columns of ``GT2'', and on submissions with at least three relevant reviewers in column ``GT3''. In ``GT1'', we report the performance without making this distinction. 

The reviewer-submission matching results of our model on the NIPS dataset are presented in Table \ref{tab:result1} alongside those of our chosen baselines.  We note that the results for APT 200 and Single Doc were only available for GT1 and we report them as such.
Some approaches including Common Topic Model, Hidden Topic Model and LDA required a hyperparameter (number of topics) to be specified. We performed experiments on NIPS data with different number of topics in Table \ref{tab:result1}. As is shown, our proposed approach consistently outperforms the strong baselines. We also note that Hidden Topic Model and Doc2Vec are competitive approaches in expertise matching compared against probabilistic models.

\begin{table}[htbp!]
\centering
\begin{tabular}{|c|c|}
\hline
\textbf{\begin{tabular}[c]{@{}c@{}}Expertise \\  Level\end{tabular}} & \textbf{\begin{tabular}[c]{@{}c@{}}\% of Reviewers \\  Predicted by the System\end{tabular}} \\ \hline
$\geq$5                                                              & 15.2                    \\ \hline
$\geq$4                                                              & 63.6                    \\ \hline
$\geq$3                                                              & 87.9                    \\ \hline
$\geq$2                                                              & 100                      \\ \hline
\end{tabular}
\caption{Percentage of reviewers in levels of expertise to the submissions recommended by our model.}
\label{tab:result2}
\end{table}

\subsection{Results on the New Dataset} 
Our proposed approach has been used to assist in the paper-reviewer matching process in a tier-1 conference of computer architecture. We evaluated our approach on a new dataset constructed with reviewers' feedbacks on their assigned submissions. Based on the optimal number of topics on the NIPS dataset, we set the number of common topics to be 10 in this experiment.

We report the percentage of reviewers whose reported expertise level falls in the given range in Table \ref{tab:result2}. We note that all recommendations made by our system are reasonable considering that all reviewers had expertise levels no lower than $2$. The majority ($87.9\%$) of reviewers reported that they were familiar with the topics of the submissions assigned to them, and $63.6\%$ of the reviewers had deep understanding of the submissions.

\begin{table*}[htbp!]
\centering
\resizebox{1.0\textwidth}{!}{
\begin{tabular}{|c|c|c|c|c|c|c|c|}
\hline
Reviewer & TREC & Research topics & CT & HT & LDA & Doc2Vec & WMD  \\ \hline
1 & 1 & \begin{tabular}[l]{@{}l@{}} Speech recognition with Bayesian approach, Neural network \end{tabular}  & 6 & 2 & 4 & 1 & 7 \\ \hline
2 & 0 & \begin{tabular}[l]{@{}c@{}} Online learning, Sequential prediction, Bayes point machine \end{tabular} & 10 & 10 & 8 & 9 & 1\\  \hline
3 & 2 & \begin{tabular}[l]{@{}c@{}} Bayesian network, Variational Bayes estimation, Mixture models\end{tabular} & 1 & 4 & 3 & 2 & 9 \\  \hline
4 & 3 & \begin{tabular}[l]{@{}c@{}} Variational method, Bayesian learning, Markov model \end{tabular} & 2 & 3 & 9 & 4 & 8 \\ \hline
5 & 3 & \begin{tabular}[l]{@{}c@{}}Bayesian learning, Variational method, Active learning\end{tabular} & 3 & 5 & 1 & 7 & 6 \\  \hline 
\end{tabular}}
\caption{Examples of reviewers and their relevance to the submission ranked by different algorithms.}
\label{tab:nipsanalsyis}
\end{table*}

\begin{table}[htbp!]
\centering
\resizebox{0.48\textwidth}{!}{
\begin{tabular}{|l|} 
\hline
\begin{tabular}[c]{@{}l@{}}Dirichlet Process (DP) mixture models are  \\
candidates for clustering applications where the \\
number of clusters is unknown a priori. [...] The \\
speedup is achieved by incorporating kd-trees \\into a
variational Bayesian algorithm for DP \\
mixture [...]
\end{tabular}  \\
\hline
\end{tabular}}
\caption{An example of abstract from a submission.}
\label{tab:submissionexample}
\end{table}

\section{Error Analysis}
We perform a qualitative analysis on NIPS dataset to analyze the difference of different algorithms on expertise matching. 
For the clarity of our discussion, we sample a submission whose abstract is shown in Table~\ref{tab:submissionexample}. We consider five models: common topic modeling (CT), hidden topic modeling (HT), LDA, Doc2Vec and WMD. We list reviewers who were considered as top candidates for this submission by the five models in Table~\ref{tab:nipsanalsyis}. For the analyses, we used research topics from the publications of the reviewers as well as their relevance scores assigned by human annotators (i.e., their TREC scores in NIPS dataset). Reviewers are sorted in decreasing order of their relevance to the submission by five models. For example, rank 1 corresponds to the highest relevance. In Table~\ref{tab:nipsanalsyis}, We also present the rank of each reviewer given by the models.

\textit{Common topic model}. According to the common topic model, reviewer 3, 4 and 5 are included as its top 3 recommendations. 
But we note that it ranks reviewer 3 higher than reviewer 4 and 5. 
The relevance scoring of common topic model is based on the relevance between the common topic ``Bayesian method'' and reviewers' profile. Since reviewer 3 is more focused on Bayesian model, it's topic-reviewer relevance is higher than reviewer 4 and 5 who have broader research interests  beyond  Bayesian  model  and  more  publications. One limitation of common topic model reflected in this case is that it does not capture the authority and experience of reviewers.

\textit{Hidden topic model}. It incorrectly considered reviewer 1 more relevant to the submission compared to reviewer 5. We note that reviewer 5 works on a broad set of research topics ranging from Bayesian model to active learning. Since the hidden topic model extracts reviewer's topics based on the topic importance without any knowledge of the submission, it is likely that Bayesian model was not selected into representative hidden topics, which results in low relevance of reviewer 5 to the given submission. 

\textit{LDA model}. LDA assigns higher relevance to reviewer 1 than reviewer 4. Reviewer 1 used Bayesian approach, whereas it was not his research focus according to his publications. 
Reviewer 4 had done extensive research in general graphical models including Bayesian model. We observed that LDA fails to capture the relevance between graph model and Bayesian model since it ignores the semantic similarity between words.

\textit{Doc2Vec model}. Doc2Vec assigned the highest relevance to reviewer 1 among all reviewers. The document representation it generates for reviewer 1's profile is similar to the representation for the submission, possibly because the key word ``Bayesian'' and ``mixture'' in the submission also occurs frequently in the profile. It suggests that Doc2Vec model might be limited to lexical overlap. 

\textit{WMD}. Reviewer 2 is included as WMD's top recommendation, whereas the research focus of reviewer 2 is sequential prediction which is irrelevant to the submission. Moreover, actually relevant reviewers 4 and 5 were excluded from WMD's top recommendations. 
This may have resulted from WMD's word-level similarity measure. 
Reviewer 2's publications had some lexical overlap with the submission (e.g., words ``Bayes'', ``algorithm'' and ``learning'', which have high frequency in the submission). WMD tends to assign high relevance scores due to such lexical overlap.
\section{Future Work}
This study used a basic version of a reviewer's profile to be the concatenation of the abstracts  from  their  previous  publications. A concrete direction for future work would be to consider enhancements in representing reviewers' profiles.  Such efforts could consider, for instance,  the temporal variation of research interests in order to capture the relevance of a given reviewer to a given topic based on the recency of the contributions to a given area.  Other efforts could involve the use of a variable number of research topics for each reviewer and exploring ways to render reviewer profiles human interpretable.  
\section{Conclusion}
We proposed an automated reviewer-paper matching algorithm via jointly finding the common research topics between submissions and reviewers' publications. Our model is based on word embeddings and efficiently captures the reviewer-paper relevance. It is robust to cases of vocabulary mismatch and partial topic overlap between submissions and reviewers -- factors that have posed problems for previous approaches. The common topic model showed strong empirical performance on a benchmark and a newly collected dataset.

\section*{Acknowledgments}
This work is supported by the IBM-ILLINOIS Center for Cognitive Computing Systems Research (C3SR) - a research collaboration as part of the IBM AI Horizons Network. We thank the EMNLP anonymous reviewers for their constructive suggestions.


\bibliography{emnlp-ijcnlp-2019}
\bibliographystyle{acl_natbib}

\end{document}